\documentclass[11pt]{article}

\usepackage[preprint]{acl}

\usepackage{times}
\usepackage{latexsym}
\usepackage[T1]{fontenc}
\usepackage[utf8]{inputenc}
\usepackage{microtype}
\usepackage{inconsolata}
\usepackage{graphicx}
\usepackage{float}
\usepackage{booktabs}
\usepackage{xcolor}
\usepackage{multirow}
\usepackage{pifont}
\usepackage{colortbl}
\usepackage{tabularx}
\usepackage{amsmath,amsfonts,amssymb,bbm}
\usepackage{algorithm}
\usepackage{algorithmic}
\usepackage{nicefrac}
\usepackage{tikz}
\usepackage{pgfplots}
\pgfplotsset{compat=1.17}
\usetikzlibrary{shapes.geometric, arrows.meta, positioning, fit, backgrounds, patterns}

\title{All-Quadrant Bounded Clipping GRPO: Closing the Unbounded Blind Spot for Stable and Generalizable Training}

\author{Chi Liu \\
  PayPal.AI \\\And
  Xin Chen \\
  PayPal.AI}

\begin{document}
\maketitle

\begin{abstract}
Group Relative Policy Optimization (GRPO) has emerged as a popular algorithm for reinforcement learning with large language models (LLMs). However, GRPO inherits PPO's token-level clipping while replacing token-level advantages with a single sequence-level advantage. Through a four-quadrant analysis of the (likelihood-ratio, advantage) space, we show that this combination leaves one quadrant---negative advantage combined with an increased likelihood ratio (Q4)---structurally unbounded, so that a few high-ratio tokens can receive very large suppressive updates that collapse entropy and narrow the reasoning boundary. To address this, we propose All-Quadrant Bounded Clipping GRPO (ABC-GRPO), which applies unconditional clipping in all four quadrants through sign-dependent boundaries. ABC-GRPO clips the likelihood ratio before multiplying by the advantage, adding a trust-region floor in Q2 and a cap in Q4---its negative-advantage branch coinciding with dual-clip PPO---to yield bounded per-step policy displacement in every quadrant. On mathematical reasoning with Qwen3 base models, ABC-GRPO attains the highest Avg@64 and Pass@64: it is statistically superior to GRPO, SAPO, and dual-clip PPO and competitive with the strongest baseline (DAPO), while maintaining substantially higher entropy; the gains transfer to MATH-500 and to out-of-domain code (HumanEval). Ablations isolate Q4 as the dominant blind spot.
\end{abstract}

\section{Introduction}
\label{sec:intro}

Reinforcement learning from human feedback~\citep{christiano2017deep,ouyang2022training} has emerged as a core technology for enhancing the reasoning capabilities of large language models (LLMs), powering recent advances in mathematical reasoning and complex problem-solving~\citep{guo2025deepseek,wei2022chain}. Group Relative Policy Optimization (GRPO)~\citep{shao2024deepseekmath} has become a popular choice for training reasoning models, eliminating the value network overhead of Proximal Policy Optimization (PPO)~\citep{schulman2017proximal,schulman2015trust} by computing advantages at the sequence level using group-relative rewards. This simplification dramatically reduces computational costs and implementation complexity.

However, while investigating ways to improve GRPO's model performance, we found that its design---specifically, retaining PPO's token-level clipping while replacing the value network with a single sequence-level advantage---is structurally incomplete. PPO's sign-dependent clipping was designed for accurate, localized token-level advantages; paired with a coarse sequence-level advantage that is applied uniformly to every token, it leaves the negative-advantage/high-ratio region unbounded, so that a few high-ratio tokens can dominate the update and suppress otherwise useful reasoning steps.

In this paper, we dissect the four quadrants of the $(r, \hat{A})$ space induced by the clipping operation and identify the one that most significantly affects model performance. Through both analysis and empirical experiments, we propose All-Quadrant Bounded Clipping GRPO (ABC-GRPO)---a GRPO-based algorithm that applies unconditional clipping in all four quadrants via sign-dependent boundaries. ABC-GRPO clips the likelihood ratio before multiplying by the advantage---coinciding with GRPO in the two quadrants it already protects and with dual-clip PPO on the negative-advantage branch---and adds a trust-region floor in Q2 and a cap in Q4; we position this as a stability (bounded per-step displacement) mechanism rather than a claim about optimizing expected reward. Our experiments on mathematical reasoning with Qwen3 base models~\citep{qwen3} show that ABC-GRPO improves both sampling accuracy (Avg@64) and exploration diversity (Pass@64): it is statistically superior to GRPO, SAPO~\citep{gao2025softadaptivepolicyoptimization}, and dual-clip PPO~\citep{ye2020mastering} and competitive with DAPO~\citep{yu2025dapoopensourcellmreinforcement}, and the gains transfer to MATH-500~\citep{lightman2023letsverify} and HumanEval~\citep{chen2021evaluating}. Critically, while standard GRPO's Pass@64 decreases during training, ABC-GRPO improves it, addressing the reasoning-boundary narrowing~\citep{yue2025rlvr} of current Reinforcement Learning with Verifiable Rewards (RLVR) methods, and it maintains substantially higher entropy throughout training.\footnote{Code: \url{https://github.com/chi2liu/ABC-GRPO}}

\section{Related Work}
\label{sec:related}

Several recent methods address GRPO's clipping and advantage-estimation limitations. DAPO~\citep{yu2025dapoopensourcellmreinforcement} introduces a ``clip-higher'' strategy with asymmetric thresholds ($\varepsilon_{\text{low}} < \varepsilon_{\text{high}}$) and token-level loss normalization, while GSPO~\citep{zheng2025groupsequencepolicyoptimization} lifts clipping from the token to the sequence level to alleviate over-punishment of individual tokens. Two methods target the negative-advantage blind spot specifically. Dual-clip PPO~\citep{ye2020mastering} adds a single lower bound $c\,\hat{A}$ (with $c>1$) on the update when $\hat{A}<0$, capping the most extreme suppressive updates but leaving the positive-advantage side untouched; our negative-advantage branch coincides with dual-clip PPO at the tight setting $c{=}1{+}\varepsilon_3$, and ABC-GRPO additionally bounds the positive-advantage (Q2) side. SAPO~\citep{gao2025softadaptivepolicyoptimization} replaces hard clipping with a smooth, temperature-controlled gate that attenuates---rather than zeroes---off-policy gradients, retaining nonzero signal at finite ratios; ABC-GRPO instead uses a hard, finite, sign-conditioned interval. We compare against DAPO, GSPO, SAPO, and dual-clip PPO in Sect.~\ref{sec:comparison}. Empirically, \citet{yue2025rlvr} document that RLVR methods \emph{narrow the reasoning boundary}---Pass@$k$ at large $k$ degrades below the base model during training; our four-quadrant analysis traces GRPO's own narrowing to its unbounded Q4 clipping. Orthogonal alignment directions such as DPO~\citep{rafailov2023direct} and REINFORCE-style methods~\citep{williams1992simple,ahmadian2024back} are complementary but not the focus of this work.

\section{Preliminaries}
\label{sec:prelim}

\paragraph{Notation.} We consider a policy $\pi_\theta$ generating responses conditioned on prompt $x$. For a response $y = (y_1, \ldots, y_T)$, we denote $r_t = \frac{\pi_\theta(y_t|x,y_{<t})}{\pi_{\text{old}}(y_t|x,y_{<t})}$ as the importance ratio at token $t$, and $A_t$ as the advantage estimate. For group-based methods sampling $G$ responses $\{y_1, \ldots, y_G\}$, we use $r_{i,t}$ and $A_i$ for the token-level ratio and sequence-level advantage of the $i$-th response, respectively. We write $\mathrm{clip}(v, a, b) = \max(\min(v, b), a)$.

\paragraph{PPO.}
PPO~\citep{schulman2017proximal} constrains policy updates via a clipped surrogate objective:
\begin{multline}
\mathcal{L}^{\text{PPO}}(\theta) = \mathbb{E}_t \Big[ \min \big( r_t A_t,\, \\
\mathrm{clip}(r_t, 1{-}\varepsilon, 1{+}\varepsilon)\, A_t \big) \Big]
\label{eq:ppo}
\end{multline}
where $\varepsilon$ is the clipping threshold (typically 0.2). The $\min$ operator creates an asymmetric trust region: when $A_t > 0$, it clips $r_t$ at $1{+}\varepsilon$; when $A_t \leq 0$, it clips at $1{-}\varepsilon$. This sign-dependent clipping assumes advantage signs are reliable, which is reasonable given PPO's token-level advantage estimation via a learned value function. We note already here that, algebraically, the $\min$ bounds the update in three of the four sign/ratio regions but still leaves the negative-advantage/high-ratio region (Q4: $A_t<0$, $r_t>1$) unbounded; PPO is only \emph{relatively} less exposed than GRPO because accurate token-level advantages make a large negative advantage on a genuinely useful token rare (Sect.~\ref{sec:motivation}).

\paragraph{GRPO.}
GRPO~\citep{shao2024deepseekmath} eliminates the value function by computing advantages at the sequence level. For each prompt $x$, it samples $G$ responses from $\pi_{\text{old}}$ and computes:
\begin{equation}
A_i = r(x, y_i) - \frac{1}{G}\sum_{j=1}^G r(x, y_j)
\label{eq:grpo_advantage}
\end{equation}
This sequence-level advantage $A_i$ is then applied uniformly to all tokens in $y_i$ using PPO's clipped objective (KL regularization omitted for brevity):
\begin{multline}
\mathcal{L}^{\text{GRPO}}(\theta) = \mathbb{E}_{x\sim\mathcal{D},\, \{y_i\} \sim \pi_{\text{old}}} \Bigg[ \frac{1}{G}\sum_{i=1}^G \frac{1}{|y_i|}\sum_{t=1}^{|y_i|} \\
\min\!\Big(r_{i,t} A_i,\, \mathrm{clip}(r_{i,t}, 1{-}\varepsilon, 1{+}\varepsilon) A_i\Big) \Bigg]
\label{eq:grpo}
\end{multline}

\section{Motivation}
\label{sec:motivation}

As formalized in Sect.~\ref{sec:prelim}, GRPO inherits PPO's clipping mechanism but replaces token-level advantages with sequence-level ones. This combination introduces several fundamental issues, as also noted by \citet{liu2025understandingr1zeroliketraining}.

First, the advantage-assignment problem. GRPO eliminates the value network by relying on a single sequence-level advantage that is applied uniformly to every token. Using a trajectory-level return to update every token is not, by itself, biased---this is exactly what REINFORCE~\citep{williams1992simple} does. The issue is statistical: the sequence-level advantage is a coarse, high-variance estimate of each token's true marginal contribution, so its signal-to-noise ratio is poor. The same intermediate token can receive advantages of opposite sign depending only on whether the final answer is correct (Fig.~\ref{fig:credit_assignment}); over an individual update this surfaces as high-variance, occasionally large, updates on tokens whose true contribution is mis-estimated. ABC-GRPO does not recover token-level advantages or resolve this attribution problem; it bounds the \emph{influence} of the noisiest such updates.

\begin{figure}[t]
\centering
\begin{tikzpicture}[
    scale=0.5,
    token/.style={draw, rounded corners=1.5pt, minimum width=0.6cm, minimum height=0.45cm, font=\tiny\ttfamily},
    token_same/.style={token, fill=blue!15},
    token_wrong/.style={token, fill=red!20, thick, draw=red!70},
    token_correct/.style={token, fill=green!20, thick, draw=green!70},
    adv_arrow/.style={-Stealth, thick, line width=1pt},
]
\node[font=\small\bfseries] at (5, 6.8) {Prompt: ``Solve: $2+3=$?''};
\node[anchor=west, font=\scriptsize\bfseries] at (0, 6.0) {Response A (Wrong):};
\node[anchor=east, font=\scriptsize, red!70!black] at (10, 6.0) {$\hat{A}_{\mathrm{seq}} = -0.5$};
\foreach \i/\tok in {0/Let, 1/'s, 2/calc-, 3/ulate, 4/:, 5/2, 6/+, 7/3} {
    \pgfmathsetmacro{\x}{0.5 + \i * 1.1}
    \node[token_same] at (\x, 5.3) {\tok};
    \draw[adv_arrow, red!70!black] (\x, 5.05) -- (\x, 4.65);
    \node[red!70!black, font=\tiny] at (\x, 4.4) {$-$};
}
\node[token_wrong] at (9.3, 5.3) {=6};
\draw[adv_arrow, red!70!black] (9.3, 5.05) -- (9.3, 4.65);
\node[red!70!black, font=\tiny] at (9.3, 4.4) {$-$};
\node[anchor=west, font=\scriptsize\bfseries] at (0, 3.5) {Response B (Correct):};
\node[anchor=east, font=\scriptsize, green!60!black] at (10, 3.5) {$\hat{A}_{\mathrm{seq}} = +0.5$};
\foreach \i/\tok in {0/Let, 1/'s, 2/calc-, 3/ulate, 4/:, 5/2, 6/+, 7/3} {
    \pgfmathsetmacro{\x}{0.5 + \i * 1.1}
    \node[token_same] at (\x, 2.8) {\tok};
    \draw[adv_arrow, green!60!black] (\x, 2.55) -- (\x, 2.15);
    \node[green!60!black, font=\tiny] at (\x, 1.9) {$+$};
}
\node[token_correct] at (9.3, 2.8) {=5};
\draw[adv_arrow, green!60!black] (9.3, 2.55) -- (9.3, 2.15);
\node[green!60!black, font=\tiny] at (9.3, 1.9) {$+$};
\node[draw=orange, thick, fill=yellow!10, rounded corners=2pt,
      text width=7.5cm, align=center, font=\scriptsize] at (5, 1.0) {
    Same tokens get opposite feedback based solely on final answer
};
\end{tikzpicture}
\caption{Coarse credit assignment in GRPO. Two responses differ only in the final token, yet the sequence-level advantage assigns opposite signs to the identical first 8 tokens---a high-variance estimate of each token's true contribution rather than a deterministic error.}
\label{fig:credit_assignment}
\end{figure}

Second, the error-reinforcement problem. A positive sequence-level advantage reinforces all preceding tokens together---including ``nonsense'' or erroneous reasoning---while a negative advantage discourages even sound earlier steps. This is tolerable for short responses but increasingly detrimental in long reasoning chains, where only a small subset of tokens determines the outcome.

Third, the over-punishment problem. GRPO inherits PPO's clipping, which leaves two sign/ratio regions unclipped:
\begin{itemize}
\item Case 1 (Q4): $A < 0$ and $r > 1+\varepsilon$
\item Case 2 (Q2): $A > 0$ and $r < 1-\varepsilon$
\end{itemize}

The two cases are not equally severe. Only Case 1 (Q4) is \emph{structurally unbounded}: because $r$ has no upper bound, a large positive ratio multiplied by a negative advantage yields an arbitrarily large suppressive update. Case 2 (Q2) is unclipped but \emph{magnitude-bounded}: since $0<r<1$ there, $|r\hat{A}| \le |\hat{A}|$, so it can bias an update's direction but not produce an unbounded one. Combined with the error-reinforcement issue, unbounded Q4 updates subject correct tokens inside a negatively advantaged sequence to excessive suppression---analogous to reward overoptimization~\citep{gao2023scaling}. This blind spot is not unique to GRPO: PPO's clipping has the same Q4 gap algebraically, but PPO is only \emph{relatively} less exposed, because its token-level value function makes a large negative advantage on a genuinely useful token rare---whereas GRPO's coarse sequence-level advantage makes exactly this case common.

The cumulative effect is a progressive narrowing of the reasoning repertoire: as valid reasoning tokens are suppressed, the policy collapses onto fewer solution strategies, ultimately degrading Pass@$k$ at large $k$ below the pre-trained base model~\citep{yue2025rlvr}. Even in the bounded Case~2, we still advocate clipping: the bound acts as entropy-style regularization~\citep{haarnoja2018soft} that improves generalization and mitigates entropy collapse during training.

\section{Method}
\label{sec:method}

\paragraph{Notation.}
We use $r_{i,t}$ to denote the importance ratio for token $t$ in sequence $i$, and $\hat{A}_i$ for the sequence-level advantage estimate. In this section, we distinguish between $A^*$, the true token-level advantage (reflecting whether a token genuinely contributes positively or negatively to the outcome), and $\hat{A}$, the estimated advantage used in GRPO, which is computed at the sequence level and uniformly applied to all tokens in that sequence. A sign error occurs when $\mathrm{sign}(\hat{A}) \neq \mathrm{sign}(A^*)$. For example, a correct token ($A^* > 0$) may receive negative feedback ($\hat{A} < 0$) simply because it appeared in a low-reward (i.e., ``failed'') sequence.

\subsection{The Four Quadrants of the $(r, \hat{A})$ Space}

To better understand how GRPO's inherited clipping mechanism behaves, we classify all possible policy update scenarios along two orthogonal dimensions: (1) the sign of the estimated advantage ($\hat{A} \lessgtr 0$), indicating whether the token should be encouraged or discouraged; and (2) the direction of policy change, captured by the likelihood ratio $r \lessgtr 1$, which reflects whether the updated policy increases or decreases the probability of the token. These two binary distinctions partition the $(r, \hat{A})$ space into four quadrants (Table~\ref{tab:four_quadrants}). As we will demonstrate, GRPO's clipping, adopted from PPO, only provides protection in two of these quadrants (Q1 and Q3); of the two unprotected quadrants, only Q4 is \emph{structurally unbounded}, while Q2 is unclipped but \emph{magnitude-bounded}.

\begin{table}[ht]
\centering
\caption{Four-quadrant analysis of GRPO's clipping. GRPO only protects Q1 and Q3. Q2 and Q4 are unclipped blind spots, but only Q4 ($\hat{A} < 0$, $r > 1$) is \emph{structurally unbounded} (a large ratio yields an unbounded suppressive update); Q2 is \emph{magnitude-bounded} ($|r\hat{A}|\le|\hat{A}|$ since $0<r<1$).}
\label{tab:four_quadrants}
\resizebox{\columnwidth}{!}{%
\begin{tabular}{ccccc}
\toprule
Quadrant & $\hat{A}$ & $r$ & GRPO Clipping Rule & Status \\
\midrule
Q1 & $> 0$ & $> 1$ & $\min(r, 1{+}\varepsilon)$ & \ding{51} Clipped \\
Q2 & $> 0$ & $< 1$ & $\min(r, 1{+}\varepsilon){=}r$ & \ding{55} Blind spot (bounded) \\
Q3 & $< 0$ & $< 1$ & $\max(r, 1{-}\varepsilon)$ & \ding{51} Clipped \\
Q4 & $< 0$ & $> 1$ & $\max(r, 1{-}\varepsilon){=}r$ & \ding{55} Blind spot (unbounded) \\
\bottomrule
\end{tabular}}%
\end{table}


\textbf{Q4} ($\hat{A} < 0$, $r > 1$) is the critical case: with $r$ unbounded above, the update can drive a token that should be encouraged toward zero probability, redistributing its mass onto irrelevant alternatives~\citep{gao2025softadaptivepolicyoptimization}. \textbf{Q2} ($\hat{A} > 0$, $r < 1$) is also unclipped but bounded: since $r\in(0,1)$, $|r\hat{A}|\le|\hat{A}|$, so it may bias an update's direction without making it unbounded.

\subsection{All-Quadrant Bounded Clipping GRPO}

Motivated by the above observation, we propose All-Quadrant Bounded Clipping GRPO (ABC-GRPO), which introduces four unconditional clipping boundaries, one for each quadrant (Fig.~\ref{fig:grpo_vs_abc}). This allows independent clipping thresholds for all four scenarios:
\begin{equation}
\tilde{r}_{i,t} = \begin{cases}
\mathrm{clip}(r_{i,t}, 1-\varepsilon_2, 1+\varepsilon_1), & \text{if } \hat{A}_i > 0 \\
\mathrm{clip}(r_{i,t}, 1-\varepsilon_4, 1+\varepsilon_3), & \text{if } \hat{A}_i \leq 0
\end{cases}
\label{eq:abc_clip}
\end{equation}
Here, $\varepsilon_1$ and $\varepsilon_2$ control the upper and lower clipping bounds for positive advantages, while $\varepsilon_3$ and $\varepsilon_4$ govern those for negative advantages. This design exposes four boundaries, three more than standard GRPO, enabling per-quadrant control over policy updates. In principle these thresholds can be set independently per quadrant; in all our experiments we fix $\varepsilon_1{=}\varepsilon_2{=}\varepsilon_3{=}\varepsilon_4{=}0.2$ (the PPO default), so the four-boundary form is a property of the framework rather than a per-task tuning knob.

Our experiments show that, although this more flexible clipping mechanism does not resolve the fundamental granularity limitation of sequence-level advantages, it effectively mitigates the adverse effects of incorrect updates through smarter, quadrant-aware clipping.
We note that GSPO~\citep{zheng2025groupsequencepolicyoptimization} takes a complementary approach by extending clipping to the sequence level (see Sect.~\ref{sec:related} for a detailed comparison).

This yields the ABC-GRPO objective. ABC-GRPO multiplies the four-boundary clipped ratio directly by the advantage (clip-before-multiply), \emph{without} PPO's outer $\min$:
\begin{multline}
\mathcal{J}_\text{ABC-GRPO}(\theta) = \mathbb{E}_{x \sim \mathcal{D},\, \{y_i\} \sim \pi_{\theta_\text{old}}} \Bigg[\frac{1}{G}\sum_{i=1}^G \frac{1}{|y_i|} \\
\sum_{t=1}^{|y_i|} \tilde{r}_{i,t}(\theta)\, \hat{A}_i \Bigg],
\label{eq:abc_objective}
\end{multline}
where $r_{i,t} = \frac{ \pi_{\theta} (y_{i,t} | x, y_{i,<t}) }{ \pi_{\theta_\text{old}} (y_{i,t} | x,y_{i,<t})}$ is the token-level importance ratio and $\hat{A}_i$ is the sequence-level advantage from GRPO.

\paragraph{Key insight.} Because the clipping acts on $r_{i,t}$ before it is multiplied by $\hat{A}_i$, the \emph{loss-level} quantity $\tilde{r}_{i,t}\hat{A}_i$ is bounded in all four quadrants. We stress that this bounds the ratio \emph{used in the loss}, not the underlying policy ratio $r_{i,t}$ itself.

\paragraph{Relation to GRPO, PPO, and dual-clip.} Clipping the ratio before multiplying by the advantage ties ABC-GRPO cleanly to prior objectives. In Q1 and Q3 the four-boundary clip reproduces PPO/GRPO's clipped update exactly, so ABC-GRPO coincides with GRPO in the two quadrants GRPO already protects; in Q2 it adds a trust-region floor and in Q4 a cap---the two regions GRPO leaves unclipped. On the negative-advantage branch ($\hat{A}_i \leq 0$), ABC-GRPO's cap at $1{+}\varepsilon_3$ coincides \emph{exactly} with dual-clip PPO~\citep{ye2020mastering} at the tight constant $c{=}1{+}\varepsilon_3$; it differs from dual-clip only on the positive-advantage branch, where it additionally imposes the Q2 floor. (Our dual-clip baseline uses the standard loose $c{=}3$, so ABC-GRPO is effectively a tighter-cap dual-clip augmented with a Q2 floor---which is why it outperforms the baseline; Sect.~\ref{sec:comparison}.) Standard GRPO is recovered by removing the Q4 cap ($\varepsilon_3 \to \infty$).

\paragraph{Why clip before multiplying (and not restore the $\min$).} We are explicit about the objective's status: $\tilde{r}_{i,t}\hat{A}_i$ is \emph{not} an estimator, lower bound, or conservative surrogate of $r_{i,t}\hat{A}_i$ (it equals GRPO in Q1/Q3 but sits \emph{above} $r_{i,t}\hat{A}_i$ in Q2/Q4). It is a trust-region/proximal objective that bounds per-step displacement in all four quadrants---the same status as dual-clip PPO, DAPO, and GSPO, none a conservative surrogate of the expected reward either. Clipping \emph{before} multiplying is necessary, not cosmetic: reinstating PPO's outer $\min$ over $\{r_{i,t}\hat{A}_i,\,\tilde{r}_{i,t}\hat{A}_i\}$ collapses ABC-GRPO back to GRPO---for $\hat{A}_i<0,\,r_{i,t}>1$ the $\min$ always reselects the raw $r_{i,t}\hat{A}_i$ (as $\tilde{r}_{i,t}\le r_{i,t}$ gives $\tilde{r}_{i,t}\hat{A}_i\ge r_{i,t}\hat{A}_i$), so no choice of $\varepsilon_1,\dots,\varepsilon_4$ lets an outer $\min$ bound Q4---and at the symmetric $\varepsilon{=}0.2$ operating point it discards the Q2 floor as well, so a min-restored four-boundary objective is identical to GRPO: its results are the GRPO row of Table~\ref{tab:main_results}.


\begin{figure}[t]
\centering
\begin{tikzpicture}[scale=0.62]

\begin{scope}[xshift=0cm]
\node[font=\small\bfseries] at (2, 4.7) {GRPO};
\fill[green!15] (2, 2) rectangle (4, 4);
\fill[red!20] (0, 2) rectangle (2, 4);
\fill[green!15] (0, 0) rectangle (2, 2);
\fill[red!25] (2, 0) rectangle (4, 2);
\draw[-Stealth, thick] (-0.2, 2) -- (4.3, 2) node[right, font=\scriptsize] {$r$};
\draw[-Stealth, thick] (2, -0.2) -- (2, 4.2) node[left, font=\scriptsize] {$\hat{A}$};
\node[below, font=\tiny] at (2, 1.85) {$1$};
\node[font=\scriptsize, align=center] at (3, 3) {Q1\\Clipped};
\node[font=\scriptsize, align=center, red!70!black] at (1, 3) {Q2\\Blind Spot};
\node[font=\scriptsize, align=center] at (1, 1) {Q3\\Clipped};
\node[font=\scriptsize, align=center, red!70!black] at (3, 1) {Q4\\Blind Spot};
\end{scope}

\begin{scope}[xshift=5.5cm]
\node[font=\small\bfseries] at (2, 4.7) {ABC-GRPO};
\fill[green!15] (2, 2) rectangle (4, 4);
\fill[green!15] (0, 2) rectangle (2, 4);
\fill[green!15] (0, 0) rectangle (2, 2);
\fill[green!15] (2, 0) rectangle (4, 2);
\draw[-Stealth, thick] (-0.2, 2) -- (4.3, 2) node[right, font=\scriptsize] {$r$};
\draw[-Stealth, thick] (2, -0.2) -- (2, 4.2) node[left, font=\scriptsize] {$\hat{A}$};
\node[below, font=\tiny] at (2, 1.85) {$1$};
\node[font=\scriptsize, align=center] at (3, 3) {Q1\\Clipped};
\node[font=\scriptsize, align=center] at (1, 3) {Q2\\Clipped};
\node[font=\scriptsize, align=center] at (1, 1) {Q3\\Clipped};
\node[font=\scriptsize, align=center] at (3, 1) {Q4\\Clipped};
\end{scope}

\end{tikzpicture}
\caption{Closing GRPO's blind spots. Left: GRPO's inherited sign-dependent clipping leaves two quadrants unprotected (red). Right: ABC-GRPO applies unconditional clipping with four independent parameters ($\varepsilon_1, \varepsilon_2, \varepsilon_3, \varepsilon_4$), closing all blind spots.}
\label{fig:grpo_vs_abc}
\end{figure}

\subsection{Gradient Behavior and Stability}

Under the ABC-GRPO objective (Eq.~\ref{eq:abc_objective}), the gradient flows through $\tilde{r}_{i,t}\hat{A}_i$; where the four-boundary clip is active, $\tilde{r}_{i,t}$ is piecewise-constant in $\theta$ and contributes no gradient. Policy updates are therefore restrained in all four quadrants---including Q4 ($\hat{A}_i\le 0$, $r_{i,t}>1$), which standard GRPO leaves unbounded.

We characterize this as a \emph{bounded-displacement} (trust-region) property, not an optimality guarantee. Under mild assumptions---bounded advantage $|\hat{A}_i|\le A_{\max}$; clipped ratio $\tilde{r}_{i,t}\in[1{-}\varepsilon_{\min},1{+}\varepsilon_{\max}]$ with $\varepsilon_{\min}{=}\min(\varepsilon_2,\varepsilon_4)$ and $\varepsilon_{\max}{=}\max(\varepsilon_1,\varepsilon_3)$; and a practical bound $G_{\max}$ on $\|\nabla_\theta\log\pi_\theta\|$---the per-token gradient is uniformly bounded,
\begin{equation}
\|\nabla_\theta \mathcal{L}_t^{\text{ABC}}\| \leq A_{\max}\,(1{+}\varepsilon_{\max})\,G_{\max},
\label{eq:bound}
\end{equation}
and zero when clipped. This bounds per-step policy displacement; like PPO's clipping it is a stability construction, not a tighter estimator of expected reward. The full derivation and the algorithm pseudocode are given in Appendices~\ref{app:gradient} and~\ref{app:algorithm}.

\section{Experiments}
\label{sec:experiments}

\subsection{Experimental Setup}
\label{sec:setup}

We train Qwen3-1.7B-Base and Qwen3-4B-Base~\citep{qwen3} on DAPO-Math-17k-Processed~\citep{yu2025dapoopensourcellmreinforcement}, sampling $G=8$ responses per prompt. ABC-GRPO uses $\varepsilon_1{=}\varepsilon_2{=}\varepsilon_3{=}\varepsilon_4{=}0.2$---the PPO default; a small sweep (Appendix~\ref{app:setup}) confirms $0.2$ is robust while $0.1$ destabilizes training, and no test-benchmark metric is used for tuning. We compare against five baselines that differ from ABC-GRPO only in their clipping mechanism: GRPO ($\varepsilon{=}0.2$), DAPO with clip-higher, GSPO ($\varepsilon{=}4{\times}10^{-4}$), SAPO~\citep{gao2025softadaptivepolicyoptimization}, and dual-clip PPO~\citep{ye2020mastering} ($c{=}3$), all sharing data and optimizer settings. We evaluate on AIME 2024, AIME 2025~\citep{aime2024_2025_hf,aime2025_opencompass}, and AMC 2023~\citep{amc23_2025_hf}, sampling 64 responses per problem and reporting Avg@64 and Pass@64~\citep{chen2021evaluating}, and we test cross-domain transfer on MATH-500~\citep{lightman2023letsverify} and HumanEval~\citep{chen2021evaluating}. \textbf{Checkpoint protocol.} All methods are evaluated at the \emph{final} checkpoint of the same pre-specified five-epoch budget; no AIME/AMC/MATH metric is used to select checkpoints or hyperparameters, and per-row run and checkpoint IDs are released with our code (a per-method training trajectory is given in Appendix~\ref{app:setup}). \textbf{Objective.} All ABC-GRPO runs optimize the single clip-before-multiply objective of Eq.~\ref{eq:abc_objective} (no outer $\min$); the same objective is used at both model scales.

\subsection{Main Results}
\label{sec:comparison}

Table~\ref{tab:main_results} compares ABC-GRPO against three baselines---GRPO, DAPO~\citep{yu2025dapoopensourcellmreinforcement}, and GSPO~\citep{zheng2025groupsequencepolicyoptimization}---across two model scales. Several patterns emerge from the results:

\begin{table*}[!t]
\centering
\caption{Comparison of clipping methods across two model scales. Avg@64 measures sampling accuracy; Pass@64 measures reasoning diversity (the reasoning boundary). $\Delta$Base shows change relative to the pre-trained base model. Best results in \textbf{bold}.}
\label{tab:main_results}
\resizebox{\linewidth}{!}{%
\begin{tabular}{llcccccccccc}
\toprule
& & \multicolumn{5}{c}{Avg@64} & \multicolumn{5}{c}{Pass@64} \\
\cmidrule(lr){3-7} \cmidrule(lr){8-12}
Scale & Method & AIME24 & AIME25 & AMC23 & Avg & $\Delta$Base & AIME24 & AIME25 & AMC23 & Avg & $\Delta$Base \\
\midrule
\multirow{5}{*}{\rotatebox{90}{\scriptsize Qwen3-1.7B}}
& Base & 3.4 & 3.2 & 24.5 & 10.4 & -- & 30.0 & 40.0 & 90.0 & 53.3 & -- \\
& GRPO & 12.3 & 5.4 & 34.7 & 17.5 & $+7.1$ & 30.0 & 26.7 & \textbf{92.5} & 49.7 & $-3.6$ \\
& DAPO & 9.3 & 6.2 & 38.9 & 18.1 & $+7.7$ & \textbf{36.7} & 36.7 & \textbf{92.5} & \textbf{55.3} & $\mathbf{+2.0}$ \\
& GSPO & 9.3 & 6.5 & \textbf{41.3} & 19.0 & $+8.6$ & 30.0 & \textbf{40.0} & \textbf{92.5} & 54.2 & $+0.9$ \\
& ABC-GRPO & \textbf{14.4} & \textbf{8.2} & 39.5 & \textbf{20.7} & $\mathbf{+10.3}$ & \textbf{36.7} & 36.7 & \textbf{92.5} & \textbf{55.3} & $\mathbf{+2.0}$ \\
\midrule
\multirow{5}{*}{\rotatebox{90}{\scriptsize Qwen3-4B}}
& Base & 8.1 & 5.9 & 37.9 & 17.3 & -- & 50.0 & 43.3 & 90.0 & 61.1 & -- \\
& GRPO & 20.3 & 20.0 & 63.1 & 34.5 & $+17.2$ & 43.3 & 40.0 & 95.0 & 59.4 & $-1.7$ \\
& DAPO & 22.6 & \textbf{21.0} & 65.9 & 36.5 & $+19.2$ & 56.7 & \textbf{56.7} & 95.0 & 69.5 & $+8.4$ \\
& GSPO & 20.7 & \textbf{21.0} & 64.9 & 35.5 & $+18.2$ & 53.3 & 43.3 & 95.0 & 63.9 & $+2.8$ \\
& ABC-GRPO & \textbf{25.9} & 20.6 & \textbf{68.4} & \textbf{38.3} & $\mathbf{+21.0}$ & \textbf{66.7} & 46.7 & \textbf{97.5} & \textbf{70.3} & $\mathbf{+9.2}$ \\
\bottomrule
\end{tabular}}%
\end{table*}

\textbf{Gains increase with $k$.} On Qwen3-4B AIME24, gains over GRPO grow from $+5.6$ at Avg@64 to $+23.4$ at Pass@64; Fig.~\ref{fig:training_dynamics}(b,c) shows ABC-GRPO's Pass@32 and Pass@64 rising well above GRPO's for most of training (peaking late) while GRPO degrades. The gap widens with $k$, indicating ABC-GRPO preserves multiple solution paths rather than collapsing to a narrow distribution.

\textbf{Robust across benchmarks and scales.} ABC-GRPO attains the highest Avg@64 at both scales (20.7 at 1.7B, 38.3 at 4B), ahead of the next-best baseline by $+1.7$ (over GSPO at 1.7B) and $+1.8$ (over DAPO at 4B). At 4B this margin over DAPO is not statistically significant (paired $p{=}0.107$), so we describe ABC-GRPO as \emph{competitive with} DAPO while \emph{significantly better than} GRPO, SAPO, and dual-clip PPO (Table~\ref{tab:variants}).

\textbf{Addressing the reasoning-boundary narrowing.} Comparing Pass@64 against the \emph{base model} (rather than GRPO) reveals a known pathology: GRPO's Pass@64 falls below base on Qwen3-1.7B (49.7 vs 53.3, $-3.6$) and Qwen3-4B (59.4 vs 61.1, $-1.7$), with AIME24 dropping $-6.7$ (43.3 vs 50.0)---confirming the \emph{reasoning-boundary narrowing} reported by \citet{yue2025rlvr}. DAPO and GSPO also recover the boundary, but by a different route (clip-higher, not Q4-bounding), so Q4-bounding is one \emph{sufficient} mechanism (isolated by our ablation, Sect.~\ref{sec:ablation}), not the only one. ABC-GRPO achieves the best joint accuracy--diversity tradeoff (highest Avg@64 and Pass@64) at both scales: at 4B, the highest Pass@64 (70.3) co-occurs with the highest Avg@64 (38.3); at 1.7B, it ties DAPO on Pass@64 (55.3) while leading on Avg@64 (20.7 vs 18.1). No single method dominates every subset: on AIME25 at 4B, DAPO/GSPO edge ABC-GRPO on Avg@64 by $0.4$ (within noise on 30 problems), consistent with an accuracy--diversity tradeoff rather than uniform superiority.

\begin{table}[t]
\centering
\caption{Trust-region baselines on Qwen3-4B. ABC-GRPO's Avg@64 gain is significant over SAPO ($+3.7$, $p{<}0.001$) and dual-clip PPO ($+3.8$, $p{=}0.001$); vs.\ DAPO (Table~\ref{tab:main_results}) the $+1.8$ gap is not significant ($p{=}0.107$). Bootstrap 95\% CIs: ABC-GRPO 38.3 [31.5,45.5] / 70.3 [62.2,78.9]; SAPO 34.6 [28.0,41.7] / 65.3 [56.7,74.2]. Entropy is steady-state training entropy.}
\label{tab:variants}
\resizebox{\columnwidth}{!}{%
\begin{tabular}{lccc}
\toprule
Method & Avg@64 & Pass@64 & Entropy \\
\midrule
GRPO & 34.5 & 59.4 & 0.036 \\
dual-clip PPO & 34.5 & 61.7 & 0.032 \\
SAPO & 34.6 & 65.3 & 0.061 \\
ABC-GRPO & \textbf{38.3} & \textbf{70.3} & \textbf{0.39} \\
\bottomrule
\end{tabular}}%
\end{table}

\subsection{Ablation Study: Quadrant Importance}
\label{sec:ablation}

To directly validate our central claim that Q4 is the critical blind spot while Q2 is mild (Sect.~\ref{sec:motivation}), we ablate each unprotected quadrant in isolation on Qwen3-4B (matching the 4B main results). ``Q$x$-only'' denotes clipping in Q1, Q3, and Q$x$ while leaving the other previously unprotected quadrant unbounded; the four-quadrant ABC-GRPO and vanilla GRPO are reported as the upper and lower references. Table~\ref{tab:ablation} reports the results.

\begin{table}[t]
\centering
\caption{Ablation on Qwen3-4B: each unprotected quadrant in isolation. ``Q$x$-only'' clips Q1, Q3, and Q$x$, leaving the other unprotected quadrant unbounded. $\Delta$ is vs.\ GRPO; Avg/Pass are averaged over AIME24/25 and AMC23. Per-benchmark breakdown in Appendix~\ref{app:extra}, Table~\ref{tab:ablation_full}.}
\label{tab:ablation}
\resizebox{\columnwidth}{!}{%
\begin{tabular}{lcccc}
\toprule
& \multicolumn{2}{c}{Avg@64} & \multicolumn{2}{c}{Pass@64} \\
\cmidrule(lr){2-3}\cmidrule(lr){4-5}
Method & Avg & $\Delta$GRPO & Avg & $\Delta$GRPO \\
\midrule
Base     & 17.3          & --              & 61.1          & --                \\
GRPO     & 34.5          & --              & 59.4          & --                \\
Q2-only  & 32.6          & $-1.9$          & 61.9          & $+2.5$            \\
Q4-only  & 37.7          & $+3.2$          & 67.2          & $+7.8$            \\
ABC-GRPO & \textbf{38.3} & $\mathbf{+3.8}$ & \textbf{70.3} & $\mathbf{+10.9}$  \\
\bottomrule
\end{tabular}}%
\end{table}

\textbf{Q4 protection recovers most of ABC-GRPO's gains.} Q4-only reaches Pass@64 $67.2$ ($+7.8$ over GRPO), recovering $\approx 72\%$ of the full four-quadrant gain ($+10.9$), and its improvement is $\approx 3\times$ Q2-only's ($+7.8$ vs $+2.5$), confirming Q4 as the dominant unbounded quadrant. On AIME24 Pass@64 the sequence Base/GRPO/Q2-only/Q4-only/ABC ($50.0/43.3/50.0/63.3/66.7$) is diagnostic: GRPO degrades below base, Q2-only only recovers to base, and Q4-only crosses it by $+13.3$.

\textbf{Q2 protection requires Q4 to be bounded first.} Q2-only's Avg@64 ($32.6$) falls $-1.9$ below GRPO: with Q4 still unbounded, its suppression of correct tokens remains the dominant pathology. Only once Q4 is bounded does Q2 clipping help---adding it to Q4-only yields a further $+2.6$ Avg@64 and $+3.4$ Pass@64 on AIME24.

\textbf{Justifying unconditional four-quadrant clipping.} Q4 protection consistently improves or matches GRPO across all three benchmarks (strict gains on AIME24 and AIME25; tied AMC23 Pass@64 with a $+4.2$ Avg@64 gain), while Q2 protection plays a smaller, complementary role whose effect varies modestly by metric and task: on AIME25, for instance, adding Q2 to Q4-only improves Pass@64 by $+3.4$ but slightly reduces Avg@64 by $-1.9$. Q4 and Q2 thus play distinct roles, motivating \emph{unconditional} clipping in both quadrants rather than GRPO's single conditional $\varepsilon$. The framework admits a distinct threshold per quadrant; we probe this degree of freedom directly (Sect.~\ref{sec:crossdomain}, Appendix~\ref{app:extra}) and find the symmetric $\varepsilon{=}0.2$ best among the configurations tried, so we report it as the operating point and do not tune per task.

\subsection{Generalization and Sensitivity}
\label{sec:crossdomain}
\begin{table}[t]
\centering
\footnotesize
\caption{Cross-domain transfer (Qwen3-4B, same terminal checkpoints as Table~\ref{tab:main_results}; no retraining/retuning/reselection). MATH-500 gains are paired-significant ($p{<}0.001$); HumanEval base pass@1/5/10${=}55.4/80.4/85.5$.}
\label{tab:crossdomain}
\begin{tabular}{lcc}
\toprule
Benchmark / metric & GRPO & ABC-GRPO \\
\midrule
MATH-500 Avg@64  & 82.6 & \textbf{84.7} \\
MATH-500 Pass@64 & 94.8 & \textbf{97.0} \\
HumanEval pass@1  & 58.1 & \textbf{60.5} \\
HumanEval pass@5  & 82.5 & \textbf{84.2} \\
HumanEval pass@10 & 87.9 & \textbf{88.7} \\
\bottomrule
\end{tabular}
\end{table}

\textbf{Cross-domain transfer.} On the \emph{same} terminal Qwen3-4B checkpoints (no retraining, retuning, or checkpoint reselection), ABC-GRPO improves both in-domain math (MATH-500~\citep{lightman2023letsverify}, $+2.1/+2.2$ Avg@64/Pass@64, both $p{<}0.001$) and out-of-domain code (HumanEval~\citep{chen2021evaluating}), confirming the rule is not benchmark-specific (Table~\ref{tab:crossdomain}). \textbf{Objective level.} Porting the rule to GSPO's sequence-level operation (ABC-GSPO, matched $\varepsilon{=}4{\times}10^{-4}$) lifts Avg@64 by $+2.2$ on Qwen3-1.7B (21.2 vs.\ 19.0); the diversity edge is $\varepsilon$-scale confounded and stated tentatively. \textbf{Sensitivity and stability.} Loosening a bound trains stably but degrades that quadrant---relaxing the Q4 cap ($\varepsilon_3{=}0.3$) collapses entropy to 0.045, while relaxing the Q2 floor ($\varepsilon_2{=}0.3$) drops Avg@64 to 30.6 (opposite sides of the accuracy--diversity frontier)---while tightening ($\varepsilon{=}0.1$ or asymmetric $\varepsilon{<}0.2$) crashes before completing. So $\varepsilon{=}0.2$ sits in a narrow stable basin: the decisive change is \emph{introducing} a finite Q4 cap (GRPO's is unbounded), not its value. We fix symmetric $\varepsilon{=}0.2$ and treat the four-boundary form as an analytical framework, not a per-task knob (full tables in Appendix~\ref{app:extra}).

\subsection{Training Dynamics and Diagnostic Analysis}

\begin{figure*}[t]
\centering
\includegraphics[width=0.56\textwidth]{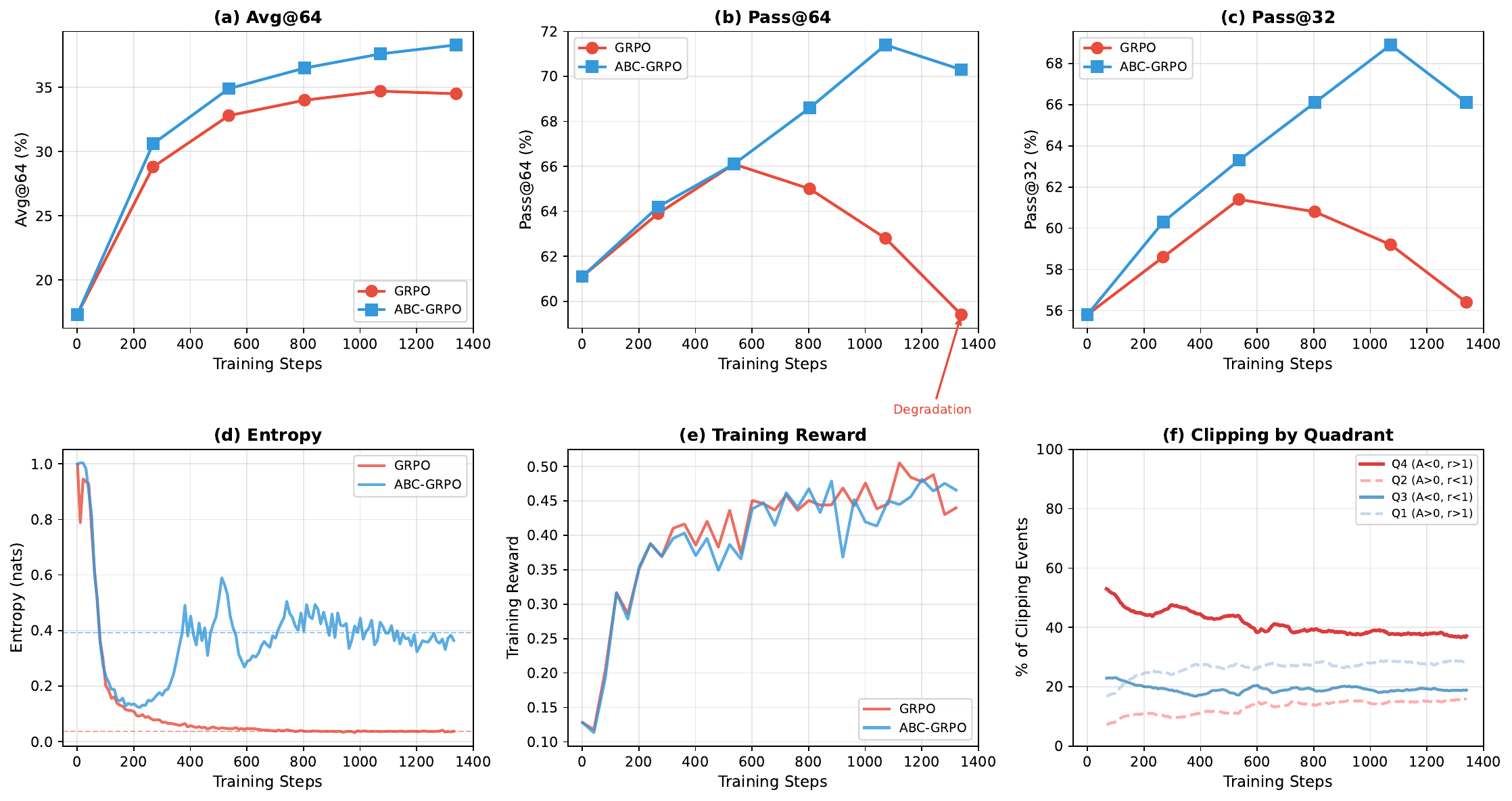}
\caption{Training dynamics for GRPO vs ABC-GRPO (Qwen3-4B). Top row shows performance metrics: (a) Avg@64 progression, (b) Pass@64 evolution showing GRPO degradation, (c) Pass@32 curves. Bottom row shows diagnostic metrics: (d) Entropy preservation, (e) Training reward curves, (f) Clipping distribution by quadrant (raw shares; Q4 is the largest category, $\approx 47\%$ after excluding zero-advantage no-op tokens).}
\label{fig:training_dynamics}
\end{figure*}

As shown in panels (a-c) of Fig.~\ref{fig:training_dynamics}, ABC-GRPO improves across Avg@64, Pass@64, and Pass@32 while GRPO degrades (Avg@64 monotonically; the Pass curves peak late, then dip slightly at the final checkpoint but stay far above GRPO). Panel (d) shows ABC-GRPO preserves substantially higher entropy throughout training: the steady-state entropy (dashed reference lines) is $\approx 0.39$ nats for ABC-GRPO versus $\approx 0.036$ nats for GRPO, a ratio of $\approx 10.8\times$ (dual-clip and SAPO likewise stay below $0.07$; Table~\ref{tab:variants}). To validate our four-quadrant analysis, panel (f) tracks clipping events during training; excluding zero-advantage no-op tokens (see Sect.~\ref{sec:clipping}), Q4 accounts for $\approx 47\%$ of all clipping events, the dominant category.

\subsection{Clipping Behavior Analysis}
\label{sec:clipping}

As shown in Fig.~\ref{fig:training_dynamics}(f), GRPO's clipping events distribute unevenly across the four quadrants of the $(r, \hat{A})$ space. We report shares after excluding zero-advantage tokens (from reward-std$=0$ groups, for which $\hat{A}{=}0$ and clipping is a no-op); the raw shares in panel (f) leave these in the denominator and therefore sum to $88.4\%$. After renormalization, the shares are Q4 $46.8\%$ (the single largest category), Q3 $25.9\%$, Q1 $19.0\%$, and Q2 $8.3\%$. The two unprotected quadrants together comprise $55.1\%$ of events, and the Q4:Q2 frequency ratio ($\approx 5.6\times$) mirrors the causal contribution ratio ($\approx 3.1\times$ on Pass@64) from the ablation in Sect.~\ref{sec:ablation}: descriptive and causal evidence converge on Q4 as the dominant blind spot.

Combined with the causal evidence from Sect.~\ref{sec:ablation}, this frequency pattern explains why bounding Q4 alone recovers most of ABC-GRPO's gains: the sequence-level/token-level mismatch (Sect.~\ref{sec:motivation}) produces a continual stream of high-ratio correct tokens inside failed sequences that GRPO's $\max(r, 1{-}\varepsilon)$ leaves unbounded, and which ABC-GRPO's $\tilde{r}_{i,t} \in [1{-}\varepsilon_4, 1{+}\varepsilon_3]$ eliminates---removing the entropy collapse of Fig.~\ref{fig:training_dynamics}(d).

\paragraph{The pathology is a tail effect.} Although the ratio is benign in aggregate (mean $1.0002$; only $0.215\%$ of tokens exceed $r{=}1.2$), it has a heavy tail (max $\approx 40.5$): a single $\hat{A}{\le}0$ token at $r{\approx}40$ receives a $\sim$$40\times$ suppressive update, and a few such events per thousand tokens over $\sim$$1.4$M token-updates per epoch suffice to collapse entropy (GRPO $0.036$ vs.\ ABC $0.39$). These events concentrate in long responses ($\geq 1024$ tokens: $64\%$ of samples but $84\%$ of Q4 events; full statistics in Appendix~\ref{app:extra}).

\section{Conclusion}

We proposed ABC-GRPO, which closes a structural gap in GRPO's clipping: a four-quadrant analysis shows only Q4 ($\hat{A}<0$, $r>1$) is structurally unbounded, and Q4 is the dominant clip category ($\approx 47\%$ of events), recovering $\approx 72\%$ of the Pass@64 gain under ablation while its heavy ratio tail collapses entropy and narrows the reasoning boundary. ABC-GRPO clips the ratio before multiplying by the advantage, adding a Q2 floor and a Q4 cap (the latter coinciding with dual-clip PPO); this bounds per-step displacement---a stability, not optimality, property. On Qwen3 base models it achieves the best accuracy--diversity tradeoff: competitive with DAPO, significantly better than GRPO, SAPO, and dual-clip PPO, with higher entropy and transfer to MATH-500 and HumanEval.

\section*{Limitations}

\paragraph{Objective justification.} The objective is a trust-region/stability construction, not a reward surrogate, so we give no optimality guarantee, and $\varepsilon{=}0.2$ is the PPO default, not a derived value.

\paragraph{Scope.} We evaluate two small same-family models (Qwen3-1.7B/4B) trained mainly on math; larger scales (7B+), other families, and other tasks remain open.

\paragraph{Credit assignment.} ABC-GRPO bounds noisy sequence-level updates but does not recover token-level advantages or disentangle other GRPO choices; configs, outputs, and checkpoint manifests are released (Sect.~\ref{sec:intro}).

\section*{Ethical Considerations}

This work improves the clipping mechanism of reinforcement learning algorithms for LLM training. It uses publicly available models and mathematical benchmarks, involves no human subjects or private data, and introduces no new risks beyond those inherent to large language model training.

\bibliography{references}

\appendix

\section{Algorithm Pseudocode}
\label{app:algorithm}

\begin{algorithm}[h]
\caption{All-Quadrant Bounded Clipping GRPO (ABC-GRPO)}
\label{alg:abc-grpo}
\begin{algorithmic}[1]
\STATE \textbf{Input:} Prompt $x$, old policy $\pi_{\text{old}}$, current policy $\pi_\theta$, clip thresholds $\{\varepsilon_1, \varepsilon_2, \varepsilon_3, \varepsilon_4\}$
\STATE Sample group of responses: $\{y_1, \ldots, y_G\} \sim \pi_{\text{old}}(\cdot|x)$
\STATE Compute rewards: $r_i = r(x, y_i)$ for $i=1,\ldots,G$
\STATE Compute advantages: $\hat{A}_i = r_i - \frac{1}{G}\sum_j r_j$
\FOR{each sequence $y_i$}
    \FOR{each token $t$ in $y_i$}
        \STATE Compute importance ratio: $r_{i,t} = \frac{\pi_\theta(y_{i,t}|x,y_{i,<t})}{\pi_{\text{old}}(y_{i,t}|x,y_{i,<t})}$
        \STATE \textbf{4-BOUNDARY CLIP:} $\tilde{r}_{i,t} \gets \begin{cases} \mathrm{clip}(r_{i,t}, 1-\varepsilon_2, 1+\varepsilon_1), & \hat{A}_i > 0 \\ \mathrm{clip}(r_{i,t}, 1-\varepsilon_4, 1+\varepsilon_3), & \hat{A}_i \leq 0 \end{cases}$
        \STATE Compute loss (clip-before-multiply): $\ell_t = -\,\tilde{r}_{i,t}\, \hat{A}_i$
    \ENDFOR
\ENDFOR
\STATE Accumulate losses and compute gradients
\end{algorithmic}
\end{algorithm}

\section{Gradient Derivation and Boundedness}
\label{app:gradient}

We give the full derivation underlying Eq.~\ref{eq:bound}. The gradient of the ABC-GRPO objective is $\nabla_\theta \mathcal{L}^{\text{ABC-GRPO}} = \mathbb{E}[\nabla_\theta \mathcal{L}_t^{\text{ABC}}]$, with per-token term $\nabla_\theta \mathcal{L}_t^{\text{ABC}} = \hat{A}_i\, \nabla_\theta \tilde{r}_{i,t}$.
Because the clipping operator is piecewise constant, its derivative is
\begin{equation}
\nabla_\theta \tilde{r}_{i,t} = \begin{cases}
\nabla_\theta r_{i,t}, & r_{i,t} \in (1{-}\varepsilon_{\text{low}}, 1{+}\varepsilon_{\text{high}}), \\
0, & \text{otherwise},
\end{cases}
\end{equation}
where $(\varepsilon_{\text{low}}, \varepsilon_{\text{high}})$ depends on the sign of $\hat{A}_i$. Using the log-derivative identity $\nabla_\theta r_{i,t} = r_{i,t}\,\nabla_\theta \log \pi_\theta(y_{i,t}|x,y_{i,<t})$ (with $\pi_{\theta_{\text{old}}}$ constant w.r.t.\ $\theta$),
\begin{equation}
\nabla_\theta \mathcal{L}_t^{\text{ABC}} = \begin{cases}
\hat{A}_i \, r_{i,t}\, \nabla_\theta \log \pi_\theta, & \text{if unclipped}, \\
0, & \text{if clipped}.
\end{cases}
\end{equation}

\paragraph{Boundedness argument.} Assume (i) $|\hat{A}_i| \leq A_{\max} < \infty$ (rewards are normalized or capped); (ii) $\tilde{r}_{i,t} \in [1{-}\varepsilon_{\min}, 1{+}\varepsilon_{\max}]$ with $\varepsilon_{\min}{=}\min(\varepsilon_2,\varepsilon_4)$ and $\varepsilon_{\max}{=}\max(\varepsilon_1,\varepsilon_3)$, so that when unclipped $r_{i,t} \leq 1{+}\varepsilon_{\max}$; and (iii) a practical bound $\|\nabla_\theta \log \pi_\theta\| \leq G_{\max}$ (finite-precision training with bounded inputs and weights). Then for any token,
\begin{multline}
\|\nabla_\theta \mathcal{L}_t^{\text{ABC}}\| \leq |\hat{A}_i|\, r_{i,t}\, \|\nabla_\theta \log \pi_\theta\| \\
\leq A_{\max}(1{+}\varepsilon_{\max})G_{\max},
\end{multline}
and zero otherwise, giving the uniform bound $C = A_{\max}(1{+}\varepsilon_{\max})G_{\max}$ of Eq.~\ref{eq:bound}.

\paragraph{Worked example (Q4 magnitude).} The per-step cap is concrete. For a Q4 token with $r{=}5,\hat{A}{=}{-}1$, GRPO/PPO applies the raw update $r\hat{A}{=}{-}5.0$; dual-clip PPO ($c{=}3$) bounds it to $-3.0$; ABC-GRPO's Q4 cap at $1{+}\varepsilon_3{=}1.2$ bounds it to $-1.2$. ABC-GRPO is thus the tightest of the three, which is why a single high-ratio negative-advantage token cannot dominate the update.

\section{Experimental Setup Details}
\label{app:setup}

\paragraph{Infrastructure and framework.}
All training runs use a single 8-GPU node with DeepSpeed ZeRO-3, mixed-precision \texttt{bfloat16}, and FlashAttention-2~\citep{dao2023flashattention2}. The trainer is a TRL-based GRPO implementation modified only to swap GRPO's clipping rule for the method-specific clipping under comparison; all other code paths, data loaders, reward functions, and logging are shared across GRPO, DAPO, GSPO, and ABC-GRPO.

\paragraph{Compute budget, software, and licenses.}
All runs use a single node of $8{\times}$H100 GPUs; each Qwen3-4B run takes $\approx$24 wall-clock hours (proportionally less for Qwen3-1.7B), and every baseline and ablation variant shares the identical compute budget at each model scale. Rollouts use vLLM~0.10.2 (colocated). Artifacts and licenses: Qwen3 base models (Apache-2.0), DAPO-Math-17k-Processed (open-source), and the AIME/AMC/MATH-500 benchmarks (publicly released for research); our released code and configurations are for research use.

\paragraph{Data and prompt template.}
We train on the open-r1/DAPO-Math-17k-Processed dataset~\citep{yu2025dapoopensourcellmreinforcement} for 5 epochs with seed $42$. Each prompt is wrapped with the system instruction \emph{``Please reason step by step, and put your final answer within}~\verb|\boxed{}|\emph{''}, matching the format used at evaluation time. The reward function is a single binary correctness signal (\texttt{accuracy}) with group-wise reward scaling enabled (\texttt{scale\_rewards=True}); no format or process reward is added.

\paragraph{Optimization.}
We use AdamW with peak learning rate $1{\times}10^{-6}$, cosine schedule, and warmup ratio $0.1$. Each device processes a per-device train batch of $4$ prompts with gradient accumulation $16$, gradient checkpointing, and \texttt{steps\_per\_generation}${=}64$, so each rollout batch is reused for $64/16{=}4$ inner optimizer passes before fresh sampling. We do not override the trainer's default KL coefficient; the same value is used for every method, isolating the clipping mechanism as the only intervention. Truncated completions are kept (\texttt{mask\_truncated\_completions=false}), and token-level importance sampling is used throughout (\texttt{importance\_sampling\_level=token}); for GSPO we additionally enable sequence-level aggregation as prescribed in the original paper~\citep{zheng2025groupsequencepolicyoptimization}.

\paragraph{Rollouts and context length.}
We sample $G{=}8$ responses per prompt via vLLM (colocated, $40\%$ GPU memory budget) with temperature $1.0$, top-$p{=}1.0$, and top-$k$ disabled. Context budgets differ slightly by model scale: Qwen3-1.7B uses (prompt $\leq 2048$, completion $\leq 4096$) and Qwen3-4B uses (prompt $\leq 1024$, completion $\leq 3072$); both budgets are shared across all baselines at the same scale. Method-specific clipping thresholds are: GRPO $\varepsilon{=}0.2$; DAPO $(\varepsilon_{\text{low}}, \varepsilon_{\text{high}}){=}(0.2, 0.28)$ following the clip-higher recipe; GSPO $\varepsilon{=}4{\times}10^{-4}$ at the sequence level; SAPO with its temperature-controlled soft gate; dual-clip PPO with $c{=}3$; and ABC-GRPO $\varepsilon_1{=}\varepsilon_2{=}\varepsilon_3{=}\varepsilon_4{=}0.2$.

\paragraph{Hyperparameter search for ABC-GRPO.}
We adopt the uniform setting at $0.2$---the community default from PPO/GRPO and dual-clip PPO---without tuning on any test benchmark. Among the configurations that trained stably, asymmetric settings (e.g., $\varepsilon_3 \neq \varepsilon_1$) did not outperform it; most other settings we attempted, including the tighter uniform $0.1$ and various asymmetric combinations, destabilized training and failed to complete. No AIME/AMC/MATH metric was used to select thresholds.

\paragraph{Evaluation protocol.}
Checkpoints are saved every epoch (4B) or every $100$ optimizer steps (1.7B) and evaluated with vLLM following the standard Qwen3-math decoding convention~\citep{qwen3}: temperature $0.6$, top-$p{=}0.95$, top-$k{=}20$, min-$p{=}0$, and decoding context length $4096$. For each problem we draw $N{=}64$ independent completions and apply the \texttt{qwen25-math-cot} answer-extraction template; Avg@64 is the mean accuracy over the $64$ samples and Pass@$k$ is the standard unbiased estimator of $\Pr[\text{any of }k\text{ samples is correct}]$~\citep{chen2021evaluating}. All reported numbers use the \emph{final} checkpoint of the fixed five-epoch budget; no test-benchmark metric is used for checkpoint or hyperparameter selection (Sect.~\ref{sec:setup}). Table~\ref{tab:trajectory} reports the Avg@64 trajectory at 20/40/60/80/100\% of training for each method, confirming that ABC-GRPO leads throughout and not merely at a cherry-picked step.

\begin{table}[H]
\centering
\caption{Qwen3-4B Avg@64 along training (fraction of the fixed five-epoch budget). All methods are compared at the same final checkpoint; ABC-GRPO leads throughout.}
\label{tab:trajectory}
\resizebox{\columnwidth}{!}{%
\begin{tabular}{lccccc}
\toprule
Method & 20\% & 40\% & 60\% & 80\% & Final \\
\midrule
GRPO & 28.8 & 32.7 & 34.1 & 34.7 & 34.5 \\
SAPO & 29.7 & 33.7 & 34.9 & 34.7 & 34.6 \\
Dual-clip & 28.0 & 33.5 & 33.4 & 34.3 & 34.5 \\
ABC-GRPO & \textbf{30.6} & \textbf{34.9} & \textbf{36.5} & \textbf{37.6} & \textbf{38.3} \\
\bottomrule
\end{tabular}}%
\end{table}

\paragraph{Benchmark details.}
AIME 2024 and AIME 2025 each contain 30 problems; AMC 2023 contains 40 problems. These complement established benchmarks such as GSM8K~\citep{cobbe2021gsm8k} and MATH~\citep{hendrycks2021math} by targeting competition-level difficulty. Pass@$k$ is computed over $k \in \{2, 4, 8, 16, 32, 64\}$; the main paper reports $k{=}64$ to track reasoning-boundary preservation, with the full $k$ sweep available in our code release.

\section{Additional Results}
\label{app:extra}

\paragraph{Per-benchmark ablation.}
Table~\ref{tab:ablation_full} gives the per-benchmark breakdown summarized in Sect.~\ref{sec:ablation}.

\begin{table}[H]
\centering
\footnotesize
\caption{Per-benchmark ablation on Qwen3-4B, expanding Table~\ref{tab:ablation}. ``Q$x$-only'' clips Q1, Q3, and Q$x$; the other unprotected quadrant remains unbounded. $\Delta$GRPO is relative to vanilla GRPO. Best per column in \textbf{bold}.}
\label{tab:ablation_full}
\resizebox{\columnwidth}{!}{%
\begin{tabular}{lccccc}
\toprule
\multicolumn{6}{c}{\emph{Avg@64}} \\
\midrule
Method & AIME24 & AIME25 & AMC23 & Avg & $\Delta$GRPO \\
\midrule
Base     & 8.1           & 5.9           & 37.9          & 17.3          & --              \\
GRPO     & 20.3          & 20.0          & 63.1          & 34.5          & --              \\
Q2-only  & 21.2          & 16.4          & 60.3          & 32.6          & $-1.9$          \\
Q4-only  & 23.3          & \textbf{22.5} & 67.3          & 37.7          & $+3.2$          \\
ABC-GRPO & \textbf{25.9} & 20.6          & \textbf{68.4} & \textbf{38.3} & $\mathbf{+3.8}$ \\
\bottomrule
\end{tabular}}%

\vspace{0.6em}

\resizebox{\columnwidth}{!}{%
\begin{tabular}{lccccc}
\toprule
\multicolumn{6}{c}{\emph{Pass@64}} \\
\midrule
Method & AIME24 & AIME25 & AMC23 & Avg & $\Delta$GRPO \\
\midrule
Base     & 50.0          & 43.3          & 90.0          & 61.1          & --                \\
GRPO     & 43.3          & 40.0          & 95.0          & 59.4          & --                \\
Q2-only  & 50.0          & 43.3          & 92.5          & 61.9          & $+2.5$            \\
Q4-only  & 63.3          & 43.3          & 95.0          & 67.2          & $+7.8$            \\
ABC-GRPO & \textbf{66.7} & \textbf{46.7} & \textbf{97.5} & \textbf{70.3} & $\mathbf{+10.9}$  \\
\bottomrule
\end{tabular}}%
\end{table}

\paragraph{Objective-level generalization (ABC-GSPO).}
Table~\ref{tab:abcgspo} reports the controlled comparison of Sect.~\ref{sec:experiments} at matched $\varepsilon{=}4{\times}10^{-4}$ on Qwen3-1.7B. Applying the four-quadrant rule to GSPO's sequence-level operation improves Avg@64 by $+2.2$. The token-level ABC configuration ($\varepsilon{=}0.2$) attains higher Pass@64/entropy, but that comparison is confounded by the $\varepsilon$-scale difference and we do not draw a strong conclusion from it.

\begin{table}[H]
\centering
\caption{Applying the four-quadrant rule at the sequence level (ABC-GSPO) on Qwen3-1.7B, at matched $\varepsilon{=}4{\times}10^{-4}$.}
\label{tab:abcgspo}
\resizebox{\columnwidth}{!}{%
\begin{tabular}{lccc}
\toprule
Method & Avg@64 & Pass@64 & Entropy \\
\midrule
GSPO & 19.0 & 54.2 & 0.120 \\
ABC-GSPO & \textbf{21.2} & 53.3 & \textbf{0.136} \\
\bottomrule
\end{tabular}}%
\end{table}

\paragraph{Threshold sensitivity.}
Table~\ref{tab:sensitivity} details the sensitivity summarized in Sect.~\ref{sec:experiments} (Qwen3-4B). Relaxing the Q4 cap collapses entropy with little reward change; relaxing the Q2 floor preserves entropy but reduces reward---the two quadrants occupy opposite sides of the accuracy--diversity frontier. Only these two single-bound relaxations (and the symmetric $0.2$ baseline) completed training; the tighter $0.1$ and the other asymmetric combinations we attempted diverged before finishing, which is why the table is sparse and we adopt the symmetric operating point.

\begin{table}[H]
\centering
\caption{Threshold sensitivity on Qwen3-4B. ``Relax'' widens one bound from the default $0.2$ to $0.3$.}
\label{tab:sensitivity}
\resizebox{\columnwidth}{!}{%
\begin{tabular}{lcc}
\toprule
Setting & Avg@64 & Entropy \\
\midrule
Default (all $\varepsilon{=}0.2$) & \textbf{38.3} & 0.39 \\
Relax Q4 ($\varepsilon_3{=}0.3$) & 33.7 & 0.045 \\
Relax Q2 ($\varepsilon_2{=}0.3$) & 30.6 & \textbf{0.776} \\
\bottomrule
\end{tabular}}%
\end{table}

\paragraph{Ratio-tail and decode-length statistics.}
Across training on Qwen3-4B, the importance ratio is benign in aggregate---mean $r{=}1.0002$, with only $0.215\%$ of tokens exceeding $r{=}1.2$---but exhibits a heavy tail, with a maximum observed ratio of $\approx 40.5$. The unbounded-Q4 pathology is therefore a tail phenomenon. The overall fraction of token-updates hitting a clip boundary is small for every method; per-method total clip ratios over training are logged in the released \texttt{trainer\_state.json} (the \texttt{clip\_ratio/*} keys). Q4 events concentrate in long responses (Table~\ref{tab:length}): responses of $\geq 1024$ tokens are $64\%$ of samples but $84\%$ of Q4 events, the 2048--3072 bucket alone accounts for $43\%$ of Q4 events, and responses under $512$ tokens contribute $\approx 1\%$; the per-token Q4 rate rises monotonically with length from $0.024\%$ to $0.036\%$.

\begin{table}[H]
\centering
\caption{Concentration of Q4 clipping events by response length (Qwen3-4B).}
\label{tab:length}
\resizebox{\columnwidth}{!}{%
\begin{tabular}{lcc}
\toprule
Response length & Share of samples & Share of Q4 events \\
\midrule
$<512$ tokens & --- & $\approx 1\%$ \\
$\geq 1024$ tokens & $64\%$ & $84\%$ \\
\quad (2048--3072) & --- & $43\%$ \\
\bottomrule
\end{tabular}}%
\end{table}


\end{document}